\documentclass[journal]{IEEETAI}                                                          % if you need a4paper
\usepackage[colorlinks,urlcolor=blue,linkcolor=blue,citecolor=blue]{hyperref}

\usepackage{color,array}

\usepackage{graphicx}

\usepackage[utf8]{inputenc}
\usepackage[T1]{fontenc}
\usepackage{times}
\usepackage{epsfig}
\usepackage{graphicx}
\usepackage{amsmath}
\usepackage{amssymb}
\usepackage{yfonts}
\usepackage{xcolor}
\usepackage{hyperref}
\usepackage[all]{hypcap}

\usepackage[noadjust]{cite}

\usepackage{fancyhdr}
\usepackage{lipsum}
\usepackage{bm,nicefrac}
\def\x{\mathbf{x}}

\begin{document}

\title{On a Sparse Shortcut Topology of Artificial Neural Networks}

\author{Feng-Lei Fan$^{1}$, \textit{Member, IEEE}, Dayang Wang$^{2}$, Hengtao Guo$^{1}$, Qikui Zhu$^{1}$, \textit{Member, IEEE}, \\ Pingkun Yan$^{1*}$, 
\textit{Senior Member, IEEE},
Ge Wang$^{1*}$, \textit{Fellow, IEEE,} and Hengyong Yu$^{2*}$, \textit{Senior Member, IEEE}% <-this % stops a space
\thanks{*Drs. Pingkun Yan, Ge Wang and Hengyong Yu serve as co-corresponding authors. This work was supported by IBM AI Horizon Scholarship, R01EB026646, R01CA233888, R01CA237267, R01HL151561, R21CA264772, and R01EB031102.}% <-this % stops a space
\thanks{$^{1}$Feng-Lei Fan (fanf2@rpi.edu), Hengtao Guo, Qikui Zhu, Pingkun Yan and Ge Wang (wangg6@rpi.edu) are with Department of Biomedical Engineering, Rensselaer Polytechnic Institute, Troy, NY 12180, USA }
\thanks{$^{2}$Dayang Wang and Hengyong Yu (hengyong\_yu@uml.edu) are with Department of Electrical and Computer Engineering, University of Massachusetts, Lowell, MA 01854, USA}
}

\markboth{IEEE Transactions on Artificial Intelligence, Vol. XX, No. X, Nov. 2021}
{Fan \MakeLowercase{\textit{et al.}}: On a Sparse Shortcut Topology of Artificial Neural Networks}

\maketitle
% \thispagestyle{empty}
% \pagestyle{empty}

%%%%%%%%%%%%%%%%%%%%%%%%%%%%%%%%%%%%%%%%%%%%%%%%%%%%%%%%%%%%%%%%%%%%%%%%%%%%%%%%
\begin{abstract}

In established network architectures, shortcut connections are often used to take the outputs of earlier layers as additional inputs to later layers. Despite the extraordinary effectiveness of shortcuts, there remain open questions on the mechanism and characteristics. For example, why are shortcuts powerful? Why do shortcuts generalize well? In this paper, we investigate the expressivity and generalizability of a novel sparse shortcut topology. First, we demonstrate that this topology can empower a one-neuron-wide deep network to approximate any univariate continuous function. Then, we present a novel width-bounded universal approximator in contrast to depth-bounded universal approximators and extend the approximation result to a family of equally competent networks. Furthermore, with generalization bound theory, we show that the proposed shortcut topology enjoys excellent generalizability. Finally, we corroborate our theoretical analyses by comparing the proposed topology with popular architectures, including ResNet and DenseNet, on well-known benchmarks and perform a saliency map analysis to interpret the proposed topology. Our work helps enhance the understanding of the role of shortcuts and suggests further opportunities to innovate neural architectures.

\end{abstract}

\begin{IEEEImpStatement}
Shortcuts are the key elements of many well-performed neural network architectures and have achieved huge success in many applications. However, over the past years, why shortcuts are powerful was not so much investigated from a theoretical point of view . To fill this gap, we present detailed analyses on the power of a sparse shortcut topology in views of expressivity and generalizability. Furthermore, our theoretical studies are corroborated by comprehensive prediction and classification experiments. Our work is useful in understanding the role of shortcuts and can inspire more research in neural architecture design.
\end{IEEEImpStatement}

\begin{IEEEkeywords}
Theoretical deep learning, network architecture, shortcut network, expressivity, generalizability
\end{IEEEkeywords}

%%%%%%%%%%%%%%%%%%%%%%%%%%%%%%%%%%%%%%%%%%%%%%%%%%%%%%%%%%%%%%%%%%%%%%%%%%%%%%%%
\section{INTRODUCTION}

Recently, deep learning \cite{lecun2015deep} has been rapidly evolving and achieved great success in many applications \cite{dahl2011context, kumar2016ask, chen2017low, wang2016perspective, anthimopoulos2016lung}. Since AlexNet \cite{krizhevsky2012imagenet}, more and more models were developed; for example, Inception \cite{szegedy2016rethinking}, Network in Network \cite{lin2013network}, VGG \cite{simonyan2014very}, ResNet \cite{he2016deep}, DenseNet \cite{huang2017densely}, and so on. These models play an important role as backbone architectures, pushing the performance boundaries of deep learning on the downstream tasks. In these studies, great efforts were made to explore the use of skip connections \cite{chu2019fast, hariharan2015hypercolumns,badrinarayanan2017segnet,srivastava2015training,larsson2016fractalnet}. For instance, a shortcut topology was searched in the framework of a lightweight network for a super-resolution task \cite{chu2019fast}. Hypercolumn Network~\cite{hariharan2015hypercolumns} stacked the units at all layers as a concatenated feature descriptor to obtain semantic information and precise localization. Highway Network \cite{srivastava2015training} achieved great success in training a very deep network. Fractal Network~\cite{larsson2016fractalnet} utilized a different skip connection design, by which interacting sub-paths were used without any pass-through or residual connections.

\begin{figure}
\centerline{\includegraphics[scale = 0.3]{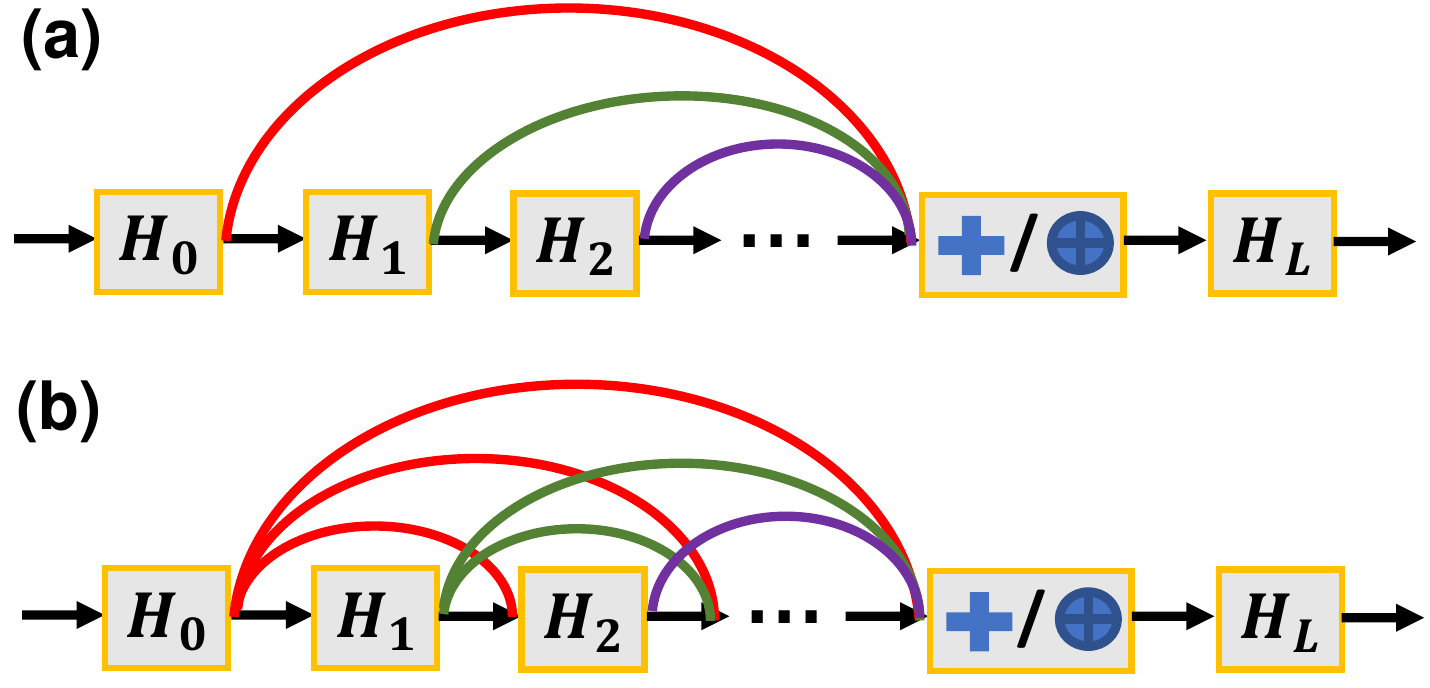}}
\caption{Comparison of sparse and dense shortcut topologies. (a) A novel sparse shortcut topology; (b) the densely connected topology, where $H_i$ denotes a collection of common operations such as convolution, ReLU, and so on. There are two aggregation methods: summation and concatenation marked as $+$ and $\oplus$, respectively. In this paper, the summation is used for expressivity and concatenation for generalization purposes.}
\label{topology}
\vspace{-5mm}
\end{figure}

In the 1990s, the universal approximation theorem was proved to justify the representation power of a network. Given a sufficient number of neurons, a one-hidden-layer network can express any continuous function \cite{funahashi1989approximate,hornik1989multilayer}. Recently, inspired by the success of deep learning, intensive efforts were put to explain the advantages of depth over width of a network. The basic idea behind these results is to construct a particular class of functions that a deep network can efficiently represent, but shallow networks cannot~\cite{szymanski2014deep,rolnick2017power,mhaskar2016deep,eldan2016power,liang2017deep}. However, despite incorporating shortcuts greatly empowers a neural network in solving real-world problems, theoretical studies are few to explain the representation and generalization abilities of shortcuts. In this study, we present our theoretical findings on a novel sparse shortcut topology, wherein shortcuts are used to bridge all prior layers and the final layer in a block or the whole network (see Figure \ref{topology}(a)), thereby partially addressing why shortcuts are effective types of machinery in a network. 

First, we show that a one-neuron-wide network with the proposed topology can approximate any univariate function, while a one-neuron-wide feedforward network cannot. This suggests that adding shortcuts can lead to a more powerful network structure. Along this direction, we report an alternative novel width-bounded universal approximator by using the Kolmogorov-Arnold representation theorem \cite{tikhomirov1991representation}, in contrast to the depth-bounded universal approximator \cite{lu2017expressive,fan2020universal,lin2018resnet}. The width-bounded universal approximator refers to the universal approximators whose width is limited, but depth is arbitrarily large, while the depth-bounded universal approximator has a limited depth, but its width can be arbitrarily large. Given the input of $n$ dimensions, the required width is no more than $2n^2+n$ per layer in our scheme. Then, we extend the result to a family of networks such that given approximation ability, these networks are equally competent. Furthermore, we analyze the effect of concatenation shortcuts on the generalization bound of deep networks. We show that the investigated topology enjoys a tighter generalization bound compared with the densely connected one, which suggests that the investigated topology can generalize well. To verify the positive results from the theoretical analyses, we prototype a network with the proposed topology and evaluate its performance on some well-known benchmarks. Finally, the experimental results demonstrate that the constructed network can achieve competitive learning performance compared to networks with residual topologies, the densely connected network, and other state-of-the-art models.

In summary, our contributions are three-fold. 1) We demonstrate the expressivity of the shortcut connections by presenting a univariate continuous function approximation theorem and a width-limited universal approximator, which partially addresses why networks with shortcuts are powerful. 2) To the best of our knowledge, our work is the first to analyze the generalizability of concatenation shortcuts based on the generalization bound theory. In addition, we also show that the generalization bounds of the proposed topology are tighter than those of the densely connected topology. 3) We conduct experiments to validate our theoretical analyses, and the investigated topology performs competitively in regression and classification experiments on several well-known benchmarks.

To clarify, all our studies are based on the architecture shown in Figure \ref{topology}(a), which is a construction of skip connections. The central hypothesis of this paper is that the proposed topology in Figure \ref{topology}(a) enjoys good expressivity (Section III) and generalizability (Section IV). Because the core of the proposed topology is the employment of shortcuts, our work also explains why shortcuts are essential in a network structure. This hypothesis is validated by comprehensive experimental comparisons (Section V).

\section{RELATED WORK}

There are studies to explain the success of summation shortcuts. It was reported in~\cite{lin2018resnet} that with residual connections, one neuron is sufficient for the ResNet to approximate any Lebesgue-integrable function. In \cite{veit2016residual}, it was showcased that the residual networks demonstrate an ensemble-like behavior. Liu \textit{et al. } \cite{liu2019towards} studied the convergence behavior of a two-layer network and proved that the optimization of a two-layer ResNet can avoid spurious minima under mild restrictions. He \textit{et al.} \cite{He2020why} studied a spectrally-normalized margin bound to discuss the influence of residual connections on the generalization ability of deep networks. They showed that the margin-based multi-class generalization bound of ResNet is of the same magnitude as that of chain-like counterparts. Therefore, the generalizability of ResNet is not worse than that of a feedforward network. Here, we not only justify the representation ability of summation shortcuts but also conduct the generalization bound analysis for concatenation shortcuts, which systematically enrich our understanding of the expressivity and generalizability of shortcuts.

The work closely related to ours was done in \cite{kang2019cycle, you2019ct}, which utilized the proposed network topology (Figure \ref{topology}(a)) as a backbone for CT image denoising and super-resolution. However, their studies were not theoretical and did not answer why such a structure can work. In contrast, we approach the utility of this shortcut topology through detailed mathematical analyses and comprehensive experiments. In addition, the investigated topology here is a sparsified version of the densely connected shortcut topology. By setting the relevant weights as zero, the densely connected topology will reduce into the topology here. Our results somehow show that the densely connected topology is redundant. 

As far as the universal approximation is concerned, in Lu \textit{et al.}~\cite{lu2017expressive}, giving at most $n+4$ neurons per layer and allowing an infinite depth, a fully-connected deep network with ReLU activation functions can accurately approximate a Lebesgue-integrable $n$-dimension function in the $L^1$-norm sense. As an extension, Lin \textit{et al.}~\cite{lin2018resnet} compressed $n+4$ into $1$ by using residual connections. They also argued that because the identity mapping should be counted as $n$ units, the actual width of their network is $n+1$. Along this direction, we exploit the Kolmogorov-Arnold representation theorem ~\cite{tikhomirov1991representation} to derive a novel width-limited universal approximator with a width no more than $2n^2+n$ per layer. Although the upper bound of width in our universal approximator is greater than those set by \cite{lu2017expressive} and \cite{lin2018resnet}, our work is still valuable because of the methodology novelty and the scarcity of width-bounded universal approximators.

\section{EXPRESSIVITY}
In this section, we first study the representation ability of the shortcut topology shown in Figure \ref{topology}(a) that is based on summation ($+$) aggregation by presenting its superior approximation ability and then extend the results to more shortcut topologies, thereby shedding light on the question why shortcuts are powerful. 

\subsection{Univariate continuous function approximation}
Our main result is that adding shortcuts, as shown in Figure \ref{topology}(a), can make a one-neuron wide network approximate any univariate continuous function in the sense of the $L^{\infty}$ distance. It should be pointed out that our result is constructive, and it is still an open problem to prove that the trained network converges to our construction. Mathematically, we make the following proposition:

\textbf{Proposition 1}: With ReLU activation functions for all hidden neurons, for any continuous function $g:[0, 1] \rightarrow \mathbb{R}$ and any given precision $\delta>0,$ there exists a neural network $G$ of the proposed topology with one neuron in each layer such that 
\begin{equation}
    \underset{x \in [0,1]}{\sup} \left | g(x) - G(x) \right | < \delta
\label{basedefine}
\end{equation} 

\textbf{The sketch of our constructive analysis}: Any univariate continuous function can be approximated by a continuous piecewise linear function within any given closeness \cite{hamann1994data}. Therefore, the key of proof becomes how to implement this piecewise approximation by a one-neuron-wide network of the proposed topology. In our scheme, we use the ReLU as activation functions for all neurons except the output neuron. By the convention of regression tasks, the activation function of the output layer is linear. Our construction is to make each neuron represent a piecewise function, and then we use shortcuts to aggregate these piecewise linear segments in the output neuron.

\textbf{Preliminaries:} Without loss of generality, a continuous function $g(x)$ can be approximated by a continuous piecewise linear function $f(x)$ at any accuracy in the $L^{\infty}$ sense, provided that the interval $[0,1]$ is partitioned into very tiny sub-intervals. Therefore, to demonstrate the correctness of \textbf{Proposition 1}, we just need to use a one-neuron-wide network of the investigated topology to implement $f(x)$. Suppose that there are $N$ pieces in $f(x)$, we can construct an explicit expression of $f(x)$ as follows: 
\begin{equation}
    f(x) =
    \begin{cases}
    f_0(x) & x\in[x_0,x_1]\\
    f_1(x) & x\in(x_1,x_2]\\
    \ \ \ \ \ \ \ \ \ \ \ \ \vdots \\
    f_{N-1}(x) & x\in(x_{N-1},x_N]
    \end{cases},
\label{fdefine}    
\end{equation}
where $x_0=0$, $x_N=1$, and 
\begin{equation}
    f_i(x) =
    \begin{cases}
   \frac{f(x_{i+1})-f(x_i)}{x_{i+1}-x_{i}}(x-x_i)+f(x_i) & x\in[x_i,x_{i+1}]\\
    0 & x\notin[x_i,x_{i+1}]\\
    \end{cases}
\label{fix}
\end{equation}
for $i = 0,1,2,\cdots,N-1$, satisfying continuity. Hereafter, we use $M_i=\frac{f(x_{i+1})-f(x_i)}{x_{i+1}-x_i}$ for simplicity. By default, neighboring segments should have different slopes; otherwise they will be combined as one segment. 

\textbf{Analysis:} Now, let us show how to select parameters of a one-neuron-wide network to express $f(x)$ in the form of Eq. \eqref{fdefine}. The outputs of neurons are respectively denoted as $R_0, R_1, R_2,...,R_{N-1} $. For the $i^{th}$ neuron, its output $R_i$ is expressed as
 \begin{equation}
\begin{aligned}
    R_i =  (W_i x + b_i)^+\label{eq},\\
\end{aligned}
\end{equation} 
where $(\cdot)^+$ denotes the ReLU operation, $W_i$ and $b_i$ are the weight and bias respectively. In the following, mathematical induction is used to show that our construction can express $f(x)$ exactly.

Initial Condition $R_0$: We use $R_0$ to implement the linear function in the first interval $[x_0, x_1]$. By setting $W_0=\left|M_0\right|,  b_0=-\left|M_0\right|x_0$, the specific function of the first neuron becomes $R_0 = \left(\left|M_0\right|(x-x_0)\right)^+$,
% \begin{equation}
%     R_0 = \left(\left|M_0\right|(x-x_0)\right)^+,
% \label{rstart}
% \end{equation}
 where the ReLU keeps the linearity when $x>x_0$.

Recurrent Relation: Suppose that we have obtained the desired $i^{th}$ neuron $R_i$, we can proceed to design the $(i+1)^{th}$ neuron with the goal of expressing the function $|f_{i+1}(x)-f_{i+1}(x_{i+1})|$, which is $|f_{i+1}(x)|$ over the interval $(x_{i+1}, x_{i+2}]$ without a constant lift. The tricky point is that the current neuron basically takes in the output of the previous neuron as the input, which is in the functional range instead of the input domain. Therefore, we need to perform an inverse affine transform:
\begin{equation}
\begin{aligned}
 &   R_{i+1} = \\
 &\left(|M_{i+1}-M_{i}|\times (\frac{1}{|M_{i}-M_{i-1}|}R_i - x_{i+1} + x_i)\right)^+
\end{aligned}
\end{equation}

For notation completeness, $M_{-1}=0$. The trick we use is to invert $R_i$ back to the input domain and set the new slope as $|M_{i+1}-M_{i}|$, which cancels the effect of $R_i$ imposed on $x>x_{i+1}$, equivalently limiting $R_i$ to only work over $(x_i,x_{i+1}]$ once $R_i$ and $R_{i+1}$ are added together. The parameters in the $(i+1)^{th}$ module are chosen as follows: $W_{i+1}=\frac{|M_{i+1}-M_{i}|}{|M_{i}-M_{i-1}|}$ and $b_{i+1}=(-x_{i+1}+x_i)|M_{i+1}-M_i|$. 

Thanks to the recurrent relation, we can compute each $R_i$ as $(|M_{i}-M_{i-1}|(x-x_i))^+$. We aggregate the outputs of those $N$ pieces in the final neuron through shortcut connections to get the neural network $G(x)$ as follows:
\begin{equation}
G(x) = \sum_{i=0}^{N-1} sgn(i)R_i + f(x_0), 
\end{equation}
wherein $sgn(i)=1$ when $M_{i}-M_{i-1}>0$ and $sgn(i)=-1$ when $M_{i}-M_{i-1}<0$. Because $R_i(x) = (|M_{i}-M_{i-1}|(x-x_i))^+$, for any $x \in [x_k,x_{k+1}]$,
\begin{equation}
\begin{aligned}
 G(x)&  = \sum_{i=0}^{N-1} sgn(i)R_i + f(x_0) \\
 & = \sum_{i=0}^{N-1} sgn(i)(|M_{i}-M_{i-1}|(x-x_i))^+ + f(x_0) \\
 & = \sum_{i=0}^{N-1} (M_{i}-M_{i-1})(x-x_i)^+ + f(x_0) \\
  & = \sum_{i=0}^{k} (M_{i}-M_{i-1})(x-x_i) + f(x_0) \\
 & = \sum_{i=0}^{k} (M_{i}-M_{i-1})x - \sum_{i=0}^{k} (M_{i}-M_{i-1})x_i + f(x_0) \\
 & = M_k x-M_k x_k + \sum_{i=0}^{k-1} M_{i}(x_{i+1}-x_i) +f(x_0) \\
 & = M_k(x-x_k) + \sum_{i=0}^{k-1} (f(x_{i+1})-f(x_i)) +f(x_0) \\
 & = M_k(x-x_k) + f(x_k) \\
 & = f_k(x),
\end{aligned}
\label{evaluation}
\end{equation}
which indicates that $G(x)$ can exactly express $f(x)$ in Eq. \eqref{fdefine}. 

\begin{figure}[htb]
\centerline{\includegraphics[scale = 0.3]{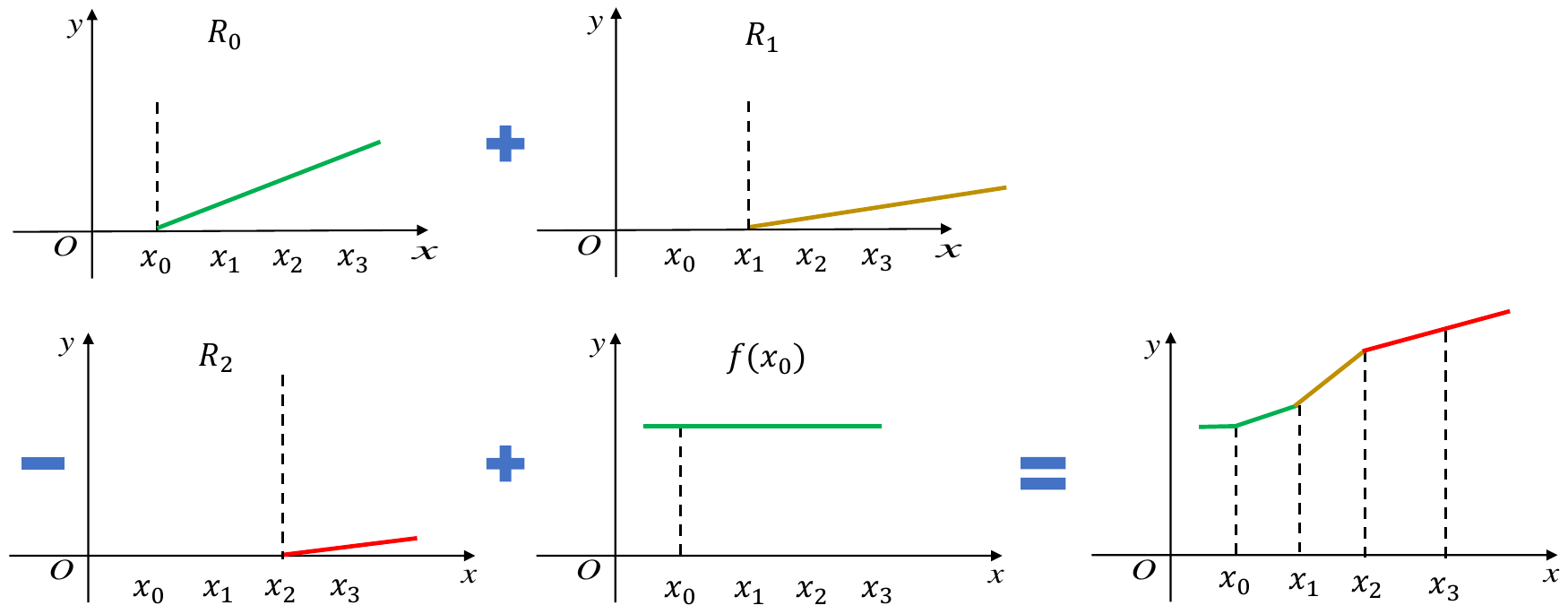}}
\caption{An example of $\sum_{i=0}^{2} sgn(i)R_i + f(x_0)$ as  $R_0+R_1-R_2+f(x_0)$ to illustrate how a one-neuron wide network can represent $f(x)$.}
\label{Slimexample}
\vspace{-5mm}
\end{figure}

To illustrate the idea clearly, we exemplify $\sum_{i=0}^{2} sgn(i)R_i + f(x_0)$ as  $R_0+R_1-R_2+f(x_0)$, as shown in Figure \ref{Slimexample}.

Based on the above derivation, for any $f(x)$ consisting of $N$ piecewise linear segments, there will be a function $f(x_0)+\sum_{i=0}^{N-1} R_i$ constructed by a one-neuron-wide $N$-layer network in the proposed topology that can exactly represent $f(x)$. Because $f(x)$ can approximate any continuous univariate function, \textbf{Proposition 1} is verified. 

Now, let us analyze the limit of $N$. Suppose $g\in C^1: [0,1]\to \mathbb{R}$, because $|g(x)-g(y)|\leq \int_{|x-y|\leq \eta} |g'(s)|ds \leq \eta ||g'||_{\infty}$, 
where $||g'||_{\infty}$ is the maximum absolute value of the derivative of $g$, a continuous piecewise linear function $f$ can represent $g$: $\sup_x |g-f| < \delta$, as long as we partition $[0,1]$ into intervals whose lengths are smaller than $\delta/||g'||_{\infty}$.  As a result, the required number of pieces is
$1/(\delta/||g'||_{\infty}) = ||g'||_{\infty}/\delta$, and the needed neuron number $N$ for $G$ is also
$||g'||_{\infty}/\delta$.

\textbf{Remark 1:} An exciting question is whether the densely connected topology in the DenseNet is necessary or not. Zhu \textit{et al.}~\cite{zhu2018sparsely} experimentally demonstrated that a sparse version of DenseNet has been excellent in image classification. In contrast, our \textbf{Proposition 1} theoretically confirms that given the sufficient depth, the densely connected topology has certain redundancy given representation ability, since a one-neuron-wide network with the proposed topology can already work for general approximation.

\subsection{Width-bounded universal approximator}

Inspired by the feasibility of using a one-neuron-wide network to approximate any continuous univariate function, here we present an alternative width-bounded universal approximator, in analogy to a depth-bounded universal approximator. Width-bounded networks mean that the width of a network is limited, but the network can be arbitrarily deep. Our scheme is based on the topology in Figure \ref{topology}(a) and the Kolmogorov-Arnold representation theorem. Specifically, we employ the Kolmogorov-Arnold representation theorem to bridge the gap between approximating univariate and multivariate functions. 

\textbf{Proposition 2}: With ReLU activation functions, for any continuous function $f:[0, 1]^n \rightarrow \mathbb{R}$ and any given precision $\sigma>0$, there exists a neural network $W$ with width no more than $2n^2+n$ per layer such that 
\begin{equation}
    \underset{x_1,x_2,...,x_n \in [0,1]}{\sup} \left | f(x_1,x_2,...,x_n) - W(x_1,x_2,...,x_n) \right | < \sigma.
\end{equation} 

\textbf{Kolmogorov-Arnold representation theorem}~\cite{tikhomirov1991representation}: For any continuous function $f(x_1,\cdots,x_n)$ with $n\geq 2$, there exist a group of continuous functions: 
$\phi_{q,p}, q = 0,1,\cdots,2n;$ $p =1,2,\cdots,n$ and $\Phi_q$ such that
\begin{equation}
    f(x_1,x_2,\cdots,x_n) = \sum_{q = 0}^{2n}\Phi_q\left(\sum_{p=1}^{n}\phi _{q,p}(x_p)\right).
\label{KATheorem}    
\end{equation}

\begin{figure}[htb]
\centerline{\includegraphics[scale = 0.26]{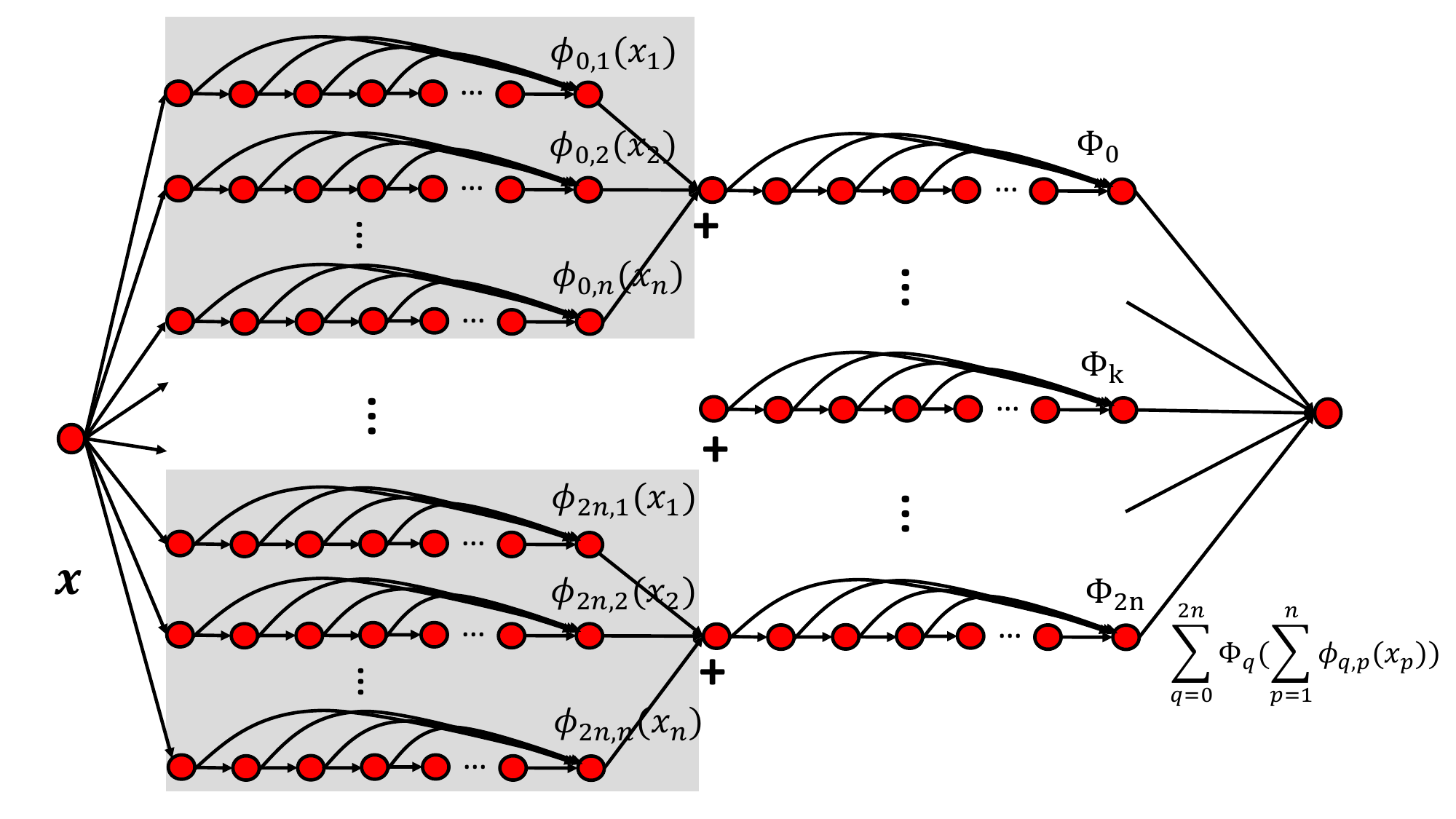}}
\caption{The scheme of a width-bounded universal approximator.}
\label{KAScheme}
\vspace{-5mm}
\end{figure}

\textbf{Scheme of analysis:} The representation theorem implies that any continuous function $f(x_1,\cdots,x_n)$ can be written as a composition of finitely many univariate functions. As shown in Figure \ref{KAScheme}, our scheme of approximating a multivariate continuous function $f(x_1, \cdots, x_n)$ is to first employ $2n^2+n$ single-neuron-wide sub-networks in the proposed topology to represent $\phi_{q,p}(x_p)$ in a parallel manner. Next, suggested by the right side of Eq. \eqref{KATheorem}, we summate the group of functions $\{\phi_{q,1}(x_1), \phi_{q,2}(x_2), ..., \phi_{q,n}(x_n)\}$ and feed $\sum_{p=1}^{n}\phi _{q,p}(x_p)$ into a new one-neuron-wide network whose purpose is to approximate $\Phi_q$. Finally, we summate the yields of those $2n+1$ sub-networks as the ultimate output of the overall network. 

\textbf{Analysis:} As we have shown in the \textbf{Proposition 1}, for every function $\phi_{q,p}(x_p)$, there exists a function $D_{q,p}(x_p)$ represented by a one-neuron-wide network in the proposed topology such that
\begin{equation}
    \underset{x_p \in [0,1]}{\sup} \left | \phi_{q,p}(x_p) - D_{q,p}(x_p) \right | < \delta_{q,p},
\label{eq_indi}    
\end{equation} 
where $\delta_{q,p}>0$ is a given arbitrarily small quantity. After we integrate $\{\phi_{q,1}(x_1), \phi_{q,2}(x_2), ..., \phi_{q,n}(x_n)\}$,
for any selection of $x_1, x_2, ..., x_n \in [0,1]$, applying triangle inequality, we obtain the error of adding $D_{q,p}$ with respect to $p$ from Eq. \eqref{eq_indi}:
\begin{equation}
\begin{aligned}
 &   \underset{x_1,x_2,...,x_n \in [0,1]}{\sup} |\sum_{p=1}^{n}\phi _{q,p}(x_p)-\sum_{p=1}^{n}D_{q,p}(x_p)|  \\
 & \leq \underset{x_1,x_2,...,x_n \in [0,1]}{\sup} \sum_{p=1}^{n}|\phi _{q,p}(x_p)-D_{q,p}(x_p)|  \\
 & < \sum_{p=1}^{n} \delta_{q,p}.
\end{aligned}
\end{equation}
Given that $\Phi_q$ is continuous, we employ the $\epsilon-\delta$ definition of continuity: if $g(x)$ is continuous at $x_0$, for any positive number $\epsilon$, there exists $\delta(\epsilon,g)>0$ satisfying that $|g(x)-g(x_0)|<\epsilon$ when $|x-x_0|<\delta$. Let $\epsilon=\frac{\sigma}{4n+2}$, correspondingly we appropriately choose $\delta_{q,p}$ so that $\sum_{p=1}^{n} \delta_{q,p} < \delta(\frac{\sigma}{4n+2},\Phi_q)$. Thus, for every $\Phi_q$, we have the following:
\begin{equation}
\begin{split}
        \underset{x_1,x_2,...,x_n \in [0,1]} {\sup}  & |\Phi_q(\sum_{p=1}^{n}\phi _{q,p}(x_p))-\Phi_q(\sum_{p=1}^{n}D_{q,p}(x_p))| \\
    & < \frac{\sigma}{4n+2}.
\end{split}
\label{continues_inequality}
\end{equation}
Every continuous function $\Phi_q$ is supported on $\mathbb{R}$ instead of $[0,1]$. Without loss of generality, we can still find a one-neuron-wide network in the proposed topology to approximate $\Phi_q$ arbitrarily well. Let $D_q(x)$ be the function expressed by such a network that can approximate $\Phi_q$ in the precision of $\frac{\sigma}{4n+2}$, we have
\begin{equation}
\begin{split}
    \underset{x \in \mathbb{R}}{\sup}  \left | \Phi_q(x) - D_q(x) \right | < \frac{\sigma}{4n+2}.
\end{split}
\label{Phi_inequality}
\end{equation}
The above equation means that $D_q(x)$ can represent $\Phi_q(x)$ with an error no greater than $\frac{\sigma}{4n+2}$ over $\mathbb{R}$. Introducing an intermediate term $\Phi_q(\sum_{p=1}^{n}D_{q,p}(x_p))$ and applying the triangle inequality to estimate the error of feeding the summation of $D_{q,p}$ into $D_q$, we have
\begin{equation}
\begin{aligned}
  & \underset{x_1,x_2,...,x_n \in [0,1]}{\sup} |\Phi_q(\sum_{p=1}^{n}\phi _{q,p}(x_p))-D_q(\sum_{p=1}^{n}D_{q,p}(x_p))| \\
  & = \underset{x_1,x_2,...,x_n \in [0,1]}{\sup} |\Phi_q(\sum_{p=1}^{n}\phi _{q,p}(x_p))-\Phi_q(\sum_{p=1}^{n}D_{q,p}(x_p)) \\
  & + \Phi_q(\sum_{p=1}^{n}D_{q,p}(x_p))-D_q(\sum_{p=1}^{n}D_{q,p}(x_p))| \\
  & \leq \underset{x_1,x_2,...,x_n \in [0,1]}{\sup} |\Phi_q(\sum_{p=1}^{n}\phi _{q,p}(x_p))-\Phi_q(\sum_{p=1}^{n}D_{q,p}(x_p))| \\
  & +  \underset{x_1,x_2,...,x_n \in [0,1]}{\sup} |\Phi_q(\sum_{p=1}^{n}D_{q,p}(x_p))-D_q(\sum_{p=1}^{n}D_{q,p}(x_p))| \\
  & < \frac{\sigma}{4n+2} + \frac{\sigma}{4n+2} = \frac{\sigma}{2n+1},
\end{aligned}
\end{equation}
where we enforce Eqs. \eqref{continues_inequality} and \eqref{Phi_inequality} to derive from the second and third lines to the fourth line. Then, applying the triangle inequality for the summation of $D_q, q=0,...,2n$, we immediately obtain the error of the total approximation scheme of Kolmogorov-Arnold  representation  theorem from Eq. \eqref{KATheorem}:
\begin{equation}
\begin{aligned}
& \underset{x_1,x_2,...,x_n \in [0,1]}{\sup}  |\sum_{q=0}^{2n} \Phi_q(\sum_{p=1}^{n}\phi _{q,p}(x_p))-\sum_{q=0}^{2n}  D_q(\sum_{p=1}^{n}D_{q,p}(x_p))| \\
& \leq \underset{x_1,x_2,...,x_n \in [0,1]}{\sup}  \sum_{q=0}^{2n} |\Phi_q(\sum_{p=1}^{n}\phi _{q,p}(x_p))- D_q(\sum_{p=1}^{n}D_{q,p}(x_p))| \\
& < (2n+1)\times\frac{\sigma}{2n+1}=\sigma.
   \end{aligned}
\end{equation}
Let $W(x_1,x_2,\cdots,x_n)=\sum_{q=0}^{2n}D_q(\sum_{p=1}^{n}D_{q,p}(x_p))$, we immediately get the validity of \textbf{Proposition 2}.  

\textbf{Remark 2:} Here, we present a novel width-limited universal approximator with a width of no more than $2n^2+n$ per layer. This width bound is greater than those of other width-bounded universal approximators, e.g., $n+4$ in \cite{lu2017expressive} and $n+1$ in \cite{lin2018resnet}. In addition, this bound is also greater than the width of common models. For example, the wide residual networks (WRN) have a width of 192, smaller than our bound. Despite that the width bound here is not pragmatic, due to the scarcity of width-bounded universal approximators and the novelty of our construction, it is still a valuable addition to the existing work.
Moreover, the Kolmogorov-Arnold representation theorem was revisited in \cite{schmidt2021kolmogorov}. The smoothness property of interior functions $\phi_{q,p}$ of the Kolmogorov-Arnold representation was enhanced by modifying the interior functions as a mapping from digits of a binary expansion to digits of a ternary expansion. Such a modification enables a ReLU network to realize the modified Kolmogorov-Arnold representation. However, the resultant network has $2K+3$ layers with $\{n, 4n,\cdots,4n,n,1,2^{Kn}+1,1\}$ neurons at each layer, respectively, where $K$ is a positive number whose value is up to the pre-specified approximation precision. Such a network is neither depth-bounded nor width-bounded.

\subsection{A family of networks}

Motivated by our constructive proof for the proposed topology, we report that in the one-dimensional setting, the aforementioned analysis is translatable to a rather inclusive family of network topologies. This network family (denoted as $\Omega^M$) subsumes an extremely wide network, an extremely deep network, and networks between them, where $M$ is the number of hidden neurons, not including the input and output nodes. We argue that network topologies in $\Omega^M$ are equivalent in the sense of the approximation ability.  

The input node is also considered as the neuron for simplicity. Hence, we refer to neurons as three types: hidden neurons, the input neuron, and the output neuron. A network in $\Omega^M$ shall satisfy the following three conditions: \\
\indent 1) Every hidden neuron has one inbound edge.  \\
\indent 2) Every hidden neuron and the input neuron have one outbound edge that links to the output neuron.  \\
\indent 3) The input neuron is wired with at least one hidden neuron. 

\begin{figure}
\center{\includegraphics[height=3.5in,width=3.5in,scale=0.35] {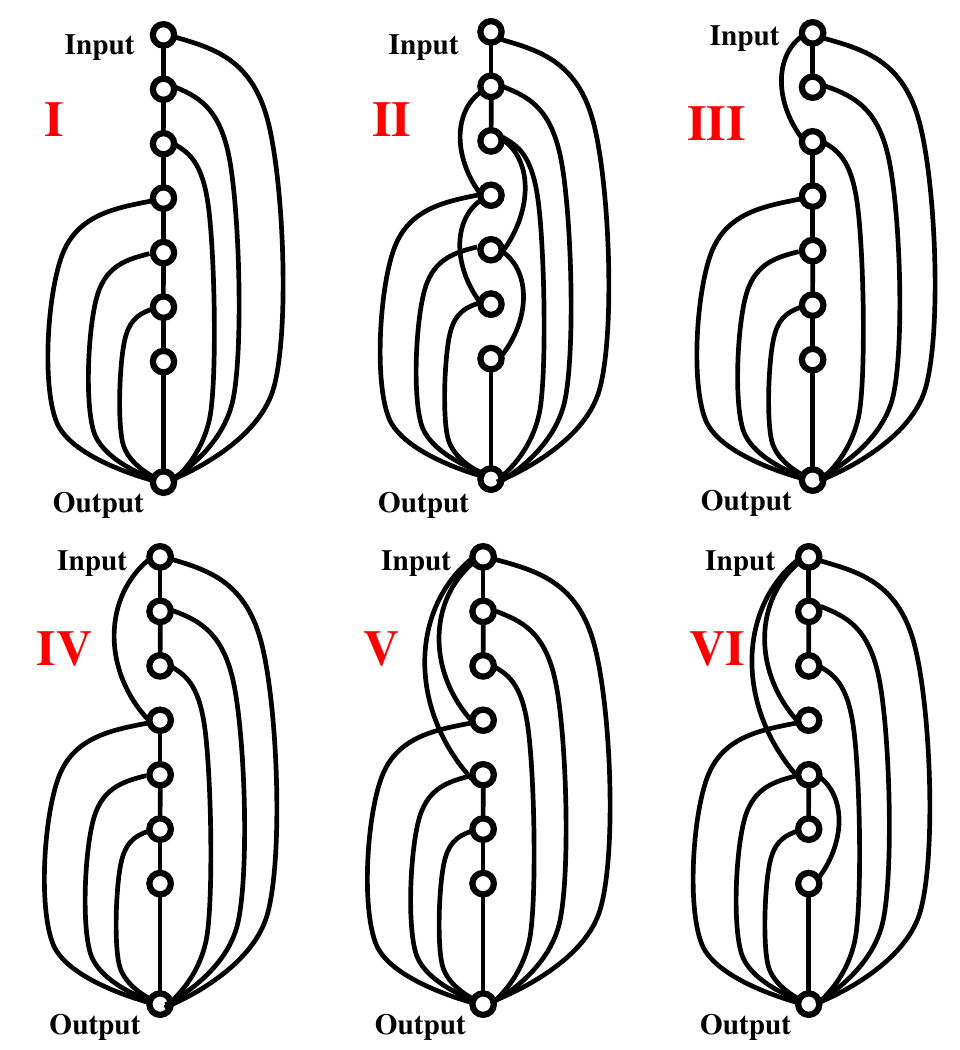}}
\caption{Six exemplary structures in $\Omega^6$ combined with ResNet setup are used to test if the networks in $\Omega$ are truly equivalent or not.}
\label{TopExample}
\vspace{-6mm}
\end{figure}

The first condition can be trivially relaxed to that every hidden neuron has multiple inbound edges by setting weights of extra edges as zero. The examples that belong to $\Omega^6$ are shown in Figure \ref{TopExample}. For a topology in $\Omega^M$, the number of required edges should be $2M+1$. One thing worthwhile to highlight is that members in $\Omega^M$ are mutually convertible through one or more cutting-rewiring operations. A cutting-rewiring process means cutting the current input edge of one neuron and rewiring the one with another neuron. Regarding the network belonging to a network family $\Omega$, we have the following proposition:  

\textbf{Proposition 3}: With ReLU activation functions, for any continuous function $g:[0, 1] \rightarrow \mathbb{R}$ and any given precision $\delta>0$, there is a network family $\Omega^{N}$ in which any network $K$, whose mapping is denoted as $\Omega_K^{N}(x)$, satisfies:
\begin{equation}
    \underset{x \in [0,1]}{\sup} \left | g(x) - \Omega_K^{N}(x) \right | < \delta.
\end{equation} 

% \begin{figure}
% \center{\includegraphics[height=2.2in,width=3.5in,scale=0.4] {Figure_2.pdf}}
% \caption{The topology of $\Omega$ comprises extremely wide and extremely deep networks}
% \end{figure}

\textbf{The sketch of analysis}: Similarly, the core of the problem is how to represent a continuous piecewise function $f(x)$ of $N$ pieces by a network from $\Omega^{N}(x)$. The main difference is that the hidden neuron in a network from $\Omega^{N}(x)$ is allowed to get the information from any previous neurons other than just precisely from the last neighboring neuron.

\textbf{Analysis:} For convenience and without loss of generality, we still use $f(x)$ in Eq. \eqref{fdefine}. To prove \textbf{Proposition 3}, we need to use $\Omega_K^{N}(x)$ to express $f(x)$.

Now we show how weights and bias in each neuron are appropriately selected in $\Omega_K^{N}(x)$ to approximate $f(x)$. Without loss of generality, the neurons are denoted as $Q_{input}, Q_0,...,Q_{N-1},Q_{output}$, where $Q_{input}$ is the input node, $Q_0$ is connected to the input neuron directly and $Q_{i+1}$ is fed with either the input neuron or another neuron $Q_t, t\leq i$, and $Q_{output}$ is the output neuron. Accordingly, the outputs of neurons $Q_0, Q_1,...,Q_{N-1}$ are also denoted as $Q_0, Q_1,...,Q_{N-1}$ for convenience, and our goal is to let $Q_0, Q_1,...,Q_{N-1}$ to represent $f_0, f_1, f_2,...,f_{N-1}$ at $[x_0,x_1], (x_1,x_2],...,(x_{N-1},x_N]$ without a constant shift. 

For $Q_0$, similar to what we did before, we set that
\begin{equation}
\begin{aligned}
    Q_0 =& \left(\left|M_0\right|(x-x_0)\right)^+.
\end{aligned}
\end{equation}

For $Q_{i+1}$, suppose that it connects with $Q_{j}$, we set 
\begin{equation}
\begin{aligned}
 &   Q_{i+1} = \\
 &\left(|M_{i+1}-M_{i}|\times (\frac{1}{|M_{j}-M_{j-1}|}Q_j - x_{i+1} + x_j)\right)^+
\end{aligned}.
\end{equation}
Thus, the output of each neuron fulfills $Q_i(x) = (|M_{i}-M_{i-1}|(x-x_i))^+$.
Similarly, we aggregate the output of all $N$ hidden neurons in the output neuron as 
\begin{equation}
\Omega_K^{N}(x)=\sum_{i=0}^{N-1} sgn(i)Q_i + f(x_0), 
\end{equation}
which is equal to $f(x)$ according to Eq. \eqref{evaluation}. Therefore, we conclude \textbf{Proposition 3}.

% \begin{table*}
% \centering
% \caption{The p-values between experimental results of all the paired networks.} 
%  \begin{tabular}{||c| c c c c c c  ||} 
%  \hline
%  &   Network I &    Network II &     Network III  &    Network IV  &       Network V&     Network VI  \\
%  \hline
%  Network I & & 0.8765 & 0.1030 & 1 & 0.2723 & 0.4743 \\
%  Network II & & & 0.0671 & 0.8738 & 0.2179 & 0.4439  \\
%  Network III & & & & 0.1007 & 0.4857 & 0.2179  \\
%  Network IV & & & & & 0.2684 & 0.4701 \\
% Network V & & & & & & 0.5856  \\
% \hline

%  \hline
%  \end{tabular}
% \end{table*}

\textbf{Remark 3:} Our representation ability analysis suggests that the members of $\Omega$ are equivalently expressive. We want to emphasize that such a finding is important in both theoretical and practical senses. On the one hand, both a one-hidden-layer but super wide network and a one-neuron-wide but super deep network are demonstrated to have a strong expressive ability. A natural curiosity is what about the networks in between. Do they also permit a good approximation ability? Here, we partially answer this question in the one-dimensional setting by showing that a wide network, a deep network, and networks in between from the network family $\Omega$ are equally capable. On the other hand, network design is an important research direction. The insight can be drawn from our finding to network architecture design and search \cite{chu2019fast}. Since many networks are actually equivalent to each other, the search and design cost will be much reduced in principle.

\section{GENERALIZATION BOUND ANALYSIS}
As mentioned earlier, for a shortcut topology, there are two types of aggregations: summation ($+$) and concatenation ($\oplus$). The effect of summation connections on the generalizability of deep networks has been studied in \cite{He2020why}. To fill the gap that the effect of concatenation shortcuts is not explored, in this section, we dissect the generalizability of concatenation shortcuts by computing the generalization bounds. A generalization bound quantifying the generalization ability of a model is the upper bound of the generalization error. Recently, aimed at explaining good generalizability of over-parameterized deep networks, a plethora of norm-based generalization bounds \cite{bartlett2017spectrally, neyshabur2015norm, neyshabur2018pac} that rely on weight matrices norms rather than the number of weights have been developed. These bounds have a better explanation because they eliminate the direct dependence on the number of parameters. 

Here, we derive the norm-based generalization bounds of DenseNet, with an emphasis on the spectrally normalized margin-based generalization bound \cite{bartlett2017spectrally}. To the best of our knowledge, our study is the first to analyze the effect of concatenation shortcuts on the generalization ability of deep networks. Then, we show that the generalization bound of the network using the proposed topology is tighter than that of the DenseNet, which suggests that the proposed topology can generalize well. 

First, the data norm is set to the $l_2$ norm and the operator norm set to the spectral norm $||\cdot||_{\sigma}$ defined as $||A||_\sigma = \underset{||Z||_2 \leq 1}{\sup} ||AZ||_2$. Furthermore, $||\cdot||_{p,q}$ is the matrix $(p,q)$-norm defined as $||A||_{p,q}=||(||A_{:,1}||_p,...,||A_{:,m_2}||_p)||_q$. 

Next, we denote the model as $F(\x)$ and define the margin operator $\mathcal{M}: \mathbb{R}^k\times \{1,2,...k\} \to \mathbb{R}$ for the $k$-class classification task as $F(\x)_z- \underset{j\neq z}{\max} F(\x)_j$ for the $z^{th}$ ground truth class, where $z$ is the class index, and the ramp function is
\begin{equation}
    l_\gamma(r) =
    \begin{cases}
    0 &  r< - \gamma\\
    1+r/\gamma &   -\gamma \leq r \leq 0\\
    1 &  r>0,\\
    \end{cases}
\end{equation}
where $\gamma$ is the margin controlling the slope of $l_\gamma(r)$. Then, the empirical ramp loss over the dataset $D=\{(\x_1,y_1),...,(\x_n,y_n)\}$ is 
\begin{equation}
\hat{\mathcal{R}}_\gamma (F) = \frac{1}{n}\sum_{i=1}^{n} (l_\gamma(-\mathcal{M}(F(\x_i),y_i))). 
\end{equation}

Minimizing the empirical ramp loss is equivalent to maximizing the margin of the predicted classes in the dataset. With all notations and definitions, we have the following theorem:

\textbf{Theorem 1:} Let us fix nonlinear activation functions $\sigma_1,\cdots,\sigma_L$, where $\sigma_i$ is $\rho_i$-Lipschitz and $\sigma_i(0)=0$. Furthermore, let the margin $\gamma > 0$, spectral norm bounds $(s_1,\cdots,s_L)$, data bound $B$, and matrices $(2,1)$-norm bounds $(b_1,\cdots,b_L)$ be given. Then, with at least $1-\delta$ probability over $N$ samples $\{(\x_i,y_i)\}_{i=1}^N$ with $\x_i \in \mathbb{R}^d,\sqrt{\sum_i ||\x_i||_2^2} \leq B$ are drawn from identical and independent distribution, every DenseNet in $F_{\mathcal{A}}: \mathbb{R}^d \to \mathbb{R}^k$ defined as 
\begin{equation}
\begin{cases}
       & G_0 = X^T \\
       &  F_1 = A_1 X^T \\
       & G_i = \sigma_i (F_i)  \\
       & F_{i+1} = A_{i+1}\oplus_{k=0}^{i}G_k \\
       & F_{L} = A_{L}\oplus_{k=0}^{L-1}G_k,
 \end{cases}
\end{equation}
where $X \in \mathbb{R}^{N\times d}$ collects all data samples $\{\x_i\}_{i=1}^N$, $\oplus$ is the matrix concatenation along the row direction, $\oplus_{k=0}^{i}G_k = G_1 \oplus G_2 \cdots \oplus G_k = [G_0;G_1;\cdots;G_k]$, $A_i$ is of $d_i \times n_i$ with  $n_i = \sum_{k=0}^{i-1}d_k$, the matrices  $\mathcal{A}=(A_1,\cdots,A_L)$ with $A_i \in \mathbb{R}^{d_i \times n_i},  n_i = \sum_{k=0}^{i-1}d_k$ obey that $||A_i||_\sigma \leq s_i$ and $||A_{i}^{T}||_{2,1}\leq b_i$, and $L$ is the number of layers, satisfies 
\begin{equation}
\begin{split}
    & Pr\{\arg\ \underset{i}{\max} F_{\mathcal{A}}(\x)_i \neq y\}-\hat{\mathcal{R}}_\gamma (F_\mathcal{A}) \\
    & \leq   \frac{8}{n^{3/2}}+3\sqrt{\frac{{\rm In}(1/\delta)}{2n}}+ \\
    & \frac{36B{\rm In}(n)\prod_{i=1}^{L}(1+\rho_i s_i)}{\gamma n}\sqrt{\sum_{i=1}^{L}\frac{\rho_i^2 b_i^2}{(1+\rho_i s_i)^2} {\rm In}(2d_i n_i) },
\end{split}
\label{GeneralizationBound}
\end{equation}
where $In(\cdot)$ is the natural logarithm.
For conciseness, we put the proof of \textbf{Theorem 1} in Part A of supplementary materials.

\textbf{Remark 4:} Please note that our result is based on the proof in \cite{bartlett2017spectrally}, and is the first to apply the results on the chain-like networks into the networks with concatenation shortcuts to evaluate the impact of concatenation shortcuts on the generalization bound of deep networks. As shown in Table \ref{comparablebounds}, we compare the bounds of the DenseNet and chain-like network. Incorporating dense concatenation shortcuts leads to a higher generalization bound than the chain-like network due to the increased matrix size $n_i=\sum_{k=0}^{i-1}d_k > d_{max}$. However, the bounds of the DenseNet and chain-like network are close when small weight matrices are used in each layer. This result partially explains why the DenseNet performs well in a small filter size because, in this situation, the concatenation shortcuts only moderately elevate the generalization bound. 

\begin{table}[htb]
 \centering

\caption{The generalization bounds for the DenseNet and chain-like network. $d_{max}$ is the maximum width. }
 \begin{tabular}{ c|c }
   \hline
      \hline
   Models & Generalization Bound \\
   \hline
    DenseNet  & $ \mathcal{O}\Big(\prod_{i=1}^{L}(1+\rho_i s_i) \sqrt{\sum_{i=1}^{L}\frac{\rho_i^2 b_i^2}{(1+\rho_i s_i)^2} {\rm In}(2d_i n_i) }\Big) $ \\
    \hline
    Chain-like & $ \mathcal{O}\Big(\prod_{i=1}^{L}(\rho_i s_i) \sqrt{\sum_{i=1}^{L}\frac{b_i^2}{ s_i^2} {\rm In}(2d_{max}^2) } \Big)$ \\
   \hline
   \hline

 \end{tabular}
 \label{comparablebounds}
\end{table}

\textbf{Proposition 4:} The margin-based multi-class generalization bound of the network in the proposed topology is tighter than that of the DenseNet. 

\textbf{Insight:} The core of the derived bound in Eq. \eqref{GeneralizationBound} is the third term of the right side, which is mainly dependent upon the spectral norm bound $s_i$ and the matrix $(2,1)$-norm bound $b_i$ of weight matrices. Because by adding imaginary shortcuts (setting the extra weight matrices as zeros), the proposed topology becomes a particular case of the DenseNet, the spectral norm bounds and matrix $(2,1)$-norm bounds of the proposed topology are no more than those of the DenseNet. Consequently, the spectrally normalized margin-based generalization bound of the network in the proposed topology is tighter than that of the DenseNet.

\textbf{Analysis:} \normalsize Let us derive the margin-based multi-class generalization bound of the network in the proposed topology and compare it with that of the DenseNet. To discriminate them, in the following we use the superscript $(S)$ for the parameters pertaining to the former and the superscript $(D)$ to the latter. Then, Eq. \eqref{GeneralizationBound} turns into 

\small
\begin{equation}
\begin{split}
    & Pr\{\arg\ \underset{i}{\max} F_{\mathcal{A}}^{(D)}(\x)_i \neq y\}-\hat{\mathcal{R}}_\gamma (F_{\mathcal{A}}^{(D)}) \\
    & \leq  \frac{8}{n^{3/2}}+3\sqrt{\frac{{\rm In}(1/\delta)}{2n}}+\\
    & \frac{36B{\rm In}(n)\prod_{i=1}^{L}(1+\rho_i s_i^{(D)})}{\gamma n}\sqrt{\sum_{i=1}^{L}\frac{\rho_i^2 b_i^{(D)2}}{{(1+\rho_i s_i^{(D)})}^2}   {\rm In}(2d_{i}n_{i}^{(D)})}.
\end{split}
\label{dense_bound}
\end{equation}
\normalsize

For a fair comparison, we set the output dimension of each layer in the network of the proposed topology to the same as that of the DenseNet. Also, we use $d_i$ for both networks. Let $A_i^{(S)} $ be of $d_i \times n_i^{(S)}$, where $ n_i^{(S)}=d_{i-1}, i\leq L-1, n_L^{(S)}=\sum_{i=1}^{L-1}d_i$ and $X \in \mathbb{R}^{n\times d}$. The computational structure of the network of the proposed topology is 
\begin{equation}
\begin{cases}
       & G_0^{(S)} = X^T \\
       &  F_1^{(S)} = A_1^{(S)} X^T  \\
       & G_i^{(S)} = \sigma_i (F_i^{(S)})  \\
       & F_{i+1}^{(S)} = A_{i+1}^{(S)} G_i^{(S)}, i\leq L-2 \\
       & F_L^{(S)} = A_{L}^{(S)} \oplus_{k=0}^{L-1}G_k^{(S)}.
 \end{cases}
\end{equation}

Without changing the final output, we can rewrite the above structure by adding imaginary shortcuts and setting the extra weight matrices as zeros, 

\begin{equation}
\begin{cases}
       & G_0^{(S)} = X^T \\
       &  F_1^{(S)} = A_1^{(S)} X^T \\
       & G_i^{(S)} = \sigma_i (F_i^{(S)})  \\
       & F_{i+1}^{(S)} = [A_{i+1}^{(S)},\textbf{0}^{d_{i+1}\times \sum_{k=0}^i d_k}][G_i^{(S)};\textbf{0}^{d_i \times n};...;\textbf{0}^{d_0 \times n}] \\
       & F_L^{(S)} = A_{L}^{(S)} \oplus_{k=0}^{L-1}G_k^{(S)},
 \end{cases}
 \label{zero_padded_densenet}
\end{equation}
where $\textbf{0}^{C_1 \times C_2}$ means the zero matrix of ${C_1 \times C_2}$. The network in the proposed topology is a special DenseNet with specific weight matrices as zeros. We can estimate the generalization bound for the above zero-padded network Eq. \eqref{zero_padded_densenet} by mimicking the generalization bound of DenseNet: 

\small
\begin{equation}
\begin{split}
    & Pr\{\arg\ \underset{i}{\max} F_{\mathcal{A}}^{(S)}(\x)_i \neq y\}-\hat{\mathcal{R}}_\gamma (F_{\mathcal{A}}^{(S)}) \\
    & \leq  \frac{8}{n^{3/2}}+3\sqrt{\frac{{\rm In}(1/\delta)}{2n}}+\\
    & \frac{36B{\rm In}(n)\prod_{i=1}^{L}(1+\rho_i s_i^{(S)})}{\gamma n}\sqrt{\sum_{i=1}^{L}\frac{\rho_i^2 b_i^{(S)2}}{(1+\rho_i s_i^{(S)})^2}   {\rm In}(2d_{i}n_{i}^{(D)})},
\end{split}
\label{proposed_bound}
\end{equation}
\normalsize
where $n_i^{(D)}$ is used because the matrix size has been enlarged to the same to that of the DenseNet.

To verify \textbf{Proposition 4}, we need to compare the bounds of DenseNet and the proposed topology (Eq. \eqref{dense_bound} vs Eq. \eqref{proposed_bound}). According to the definition of the spectral norm, we have
\begin{equation}
\begin{split}
    & ||A_i^{(S)}||_\sigma  \\
     = & \underset{||Z||_2 \leq 1}{{\rm sup}} ||A_i^{(S)}Z^{(S)}||_2 \\
     = &\underset{||Z||_2 \leq 1}{{\rm sup}} ||[A_i^{(S)},\textbf{0}^{d_i \times (n_i^{(D)}-n_i^{(S)})}]||[Z;\textbf{0}^{(n_i^{(D)}-n_i^{(S)})\times n}]||_2 \\
     \leq & \underset{||Z||_2 \leq 1}{{\rm sup}}  ||A_i^{(D)}Z^{(D)}||_2 \\
     = & ||A_i^{(D)}||_\sigma,
\end{split}
\end{equation}
where zero padding is to make $[A_i^{(S)},\textbf{0}]$ have the same size as that of $A_i^{(D)}$. Therefore, we derive that 
\begin{equation}
 s_i^{(S)} \leq s_i^{(D)}, i=1,...,L.   
\label{norm1} 
\end{equation}
In the same spirit, we can also derive that
\begin{equation}
 b_i^{(S)} \leq b_i^{(D)}, i=1,...,L. 
\label{norm2}  
\end{equation}
Combining Eqs. \eqref{norm1} and \eqref{norm2}, we have
\begin{equation}
\begin{split}
        & \prod_{i=1}^{L}(1+\rho_i s_i^{(S)})\sqrt{\sum_{i=1}^{L}\frac{\rho_i^2 b_i^{(S)2}}{(1+\rho_i^2 s_i^{(S)})^2}  {\rm In}(2d_{i}n_{i}^{(D)})}\leq \\
        & \prod_{i=1}^{L}(1+\rho_i s_i^{(D)})\sqrt{\sum_{i=1}^{L}\frac{\rho_i^2 b_i^{(D)2}}{(1+\rho_i s_i^{(D)})^2} {\rm In}(2d_{i}n_{i}^{(D)})},
\end{split}
\end{equation}
which has validated \textbf{Proposition 4}. 

\textbf{Remark 5:} Our representation and generalization analyses suggest that DenseNet has certain redundancy in representation ability and a higher generalization bound. However, the redundant structure of DenseNet may facilitate the over-parameterization effect, which may cause optimization and generalization merits. For instance, regarding merits in optimization, stochastic gradient descent (SGD) can find the global minimum in shallow or deep networks in the setting of over-parameterization because there is a large set of global minimizers in an overly parameterized network \cite{wu2018sgd, allen2018learning, brutzkus2018sgd}. Over-parameterization is also beneficial for generalization \cite{neyshabur2018role, nakkiran2019deep}. Recently, the deep double descent phenomenon (When the model complexity increases, the generalization error goes down first and then up. However, as the model complexity keeps increasing and surpasses the so-called "interpolation threshold", the generalization error starts going down) has been widely observed in many deep models \cite{nakkiran2019deep}. In light of the double descent phenomenon, the complexity of DenseNet likely lies beyond the interpolation threshold.

\section{EXPERIMENTS}

In this section, we conduct prediction and classification experiments on well-known benchmarks to evaluate the expressivity, generalizability, and interpretability of the proposed topology. The expressivity experiments use summation ($+$) shortcuts, while other experiments use concatenation ($\oplus$) shortcuts. The competitive performance on prediction and classification tasks shows that the proposed topology is a desirable architecture, as suggested by encouraging theoretical analyses. In addition, we also demonstrate the superior interpretability of the investigated topology given the saliency map.

\subsection{Expressivity}
\begin{figure}[htb]
\centerline{\includegraphics[scale = 0.4]{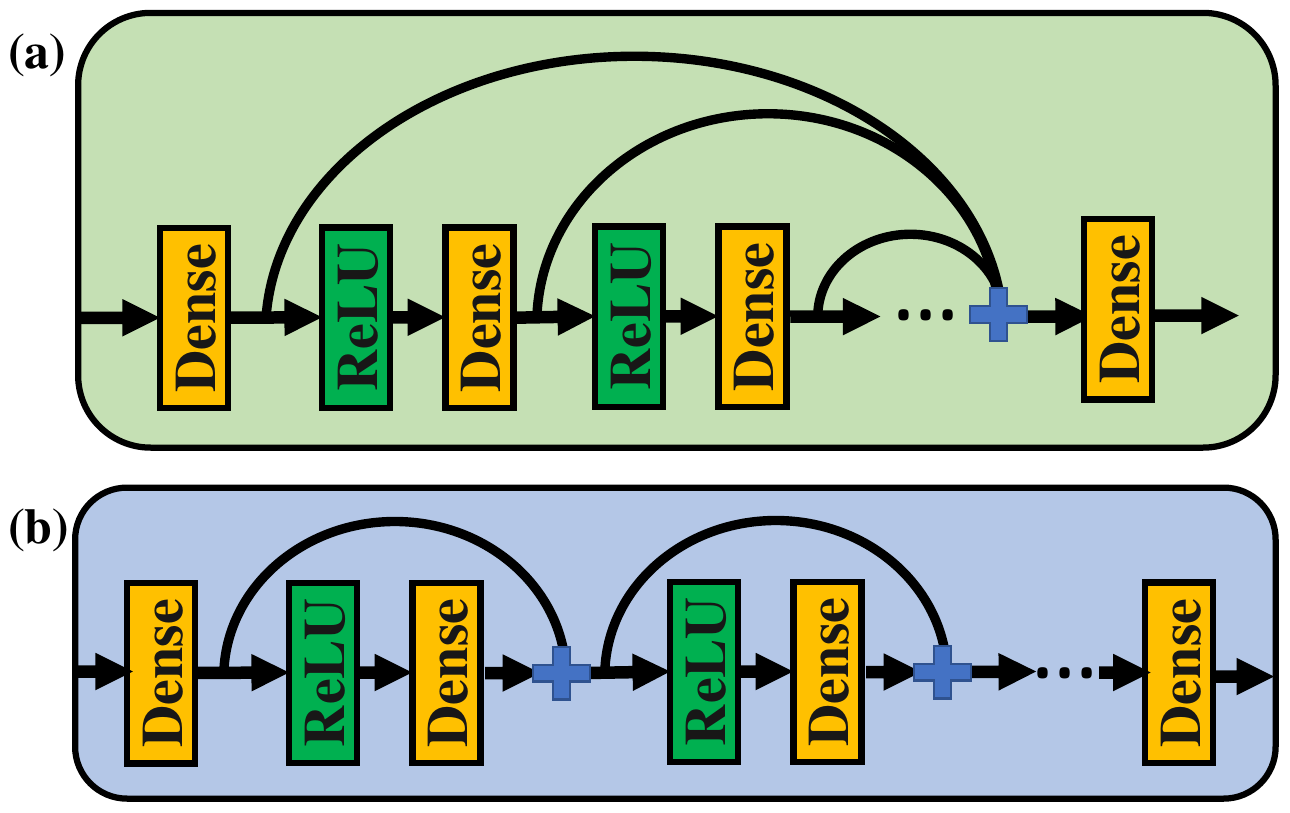}}
\caption{"Dense" denotes a fully connected layer. (a) The network of the proposed topology; (b) the network of residual topology. }
\label{NTK_topology}

\end{figure}

We compare the expressivity of the proposed topology and residual topology in the infinite-width limit, where the gradient descent makes little change to the weights of a network. The training of a neural network with infinite width in each layer turns into a kernel ridge regression \cite{vovk2013kernel} process with the so-called neural tangent kernel (NTK \cite{jacot2018neural}). When one fixes the type of activation functions, the neural tangent kernel of a neural network is only determined by the topology and the depth of the network \cite{arora2019exact}. Figure \ref{NTK_topology} shows the structures of the proposed network and a residual network that uses pre-activation features. In the proposed network, the output of each dense layer is connected to a layer before the final dense layer for summation. We denote the depth of both networks as $K+2$, where $K$ is the number of residual blocks or the number of layers that constitute the proposed topology. Two networks are the same except for shortcut architectures.

Let samples of the training dataset be $\{(\bm{x}_i, y_i)\}_{i=1}$, where $\bm{x}_i$ is the input and $y_i$ is the output, and assume that $f(\bm{\theta}, x)$ denotes the output of a neural network, where $\bm{\theta}$ are all parameters, the $(i,j)$-entry of the NTK kernel $\bm{H}^*$ \cite{arora2019exact} is defined by 
\begin{equation}
    ker(\bm{x}_i, \bm{x}_j) = \underset{\bm{\theta} \sim \bm{\Theta}}{\mathbb{E}} \left< \frac{\partial{f(\bm{\theta}, \bm{x}_i)}}{\partial \bm{\theta}}, \frac{\partial{f(\bm{\theta}, \bm{x}_j)}}{\partial \bm{\theta}} \right>.
\end{equation}
The inference process is deterministic: 
\begin{equation}
    f(\bm{x}) = [ker(\bm{x},\bm{x}_1),...,ker(\bm{x},\bm{x}_n)]\cdot (\bm{H}^*)^{-1}\bm{y}.
\end{equation}
Since the kernel in the inference process is only determined by the topology, depth, and the activation function, the comparison in the NTK domain can avoid the impact of other hyper-parameters such as the network width and learning rules (learning rate, batch size, optimizer, epoch number, and so on), which helps reveal the difference in the representation ability between two topologies.

\begin{figure}[htb]
\centerline{\includegraphics[scale = 0.6]{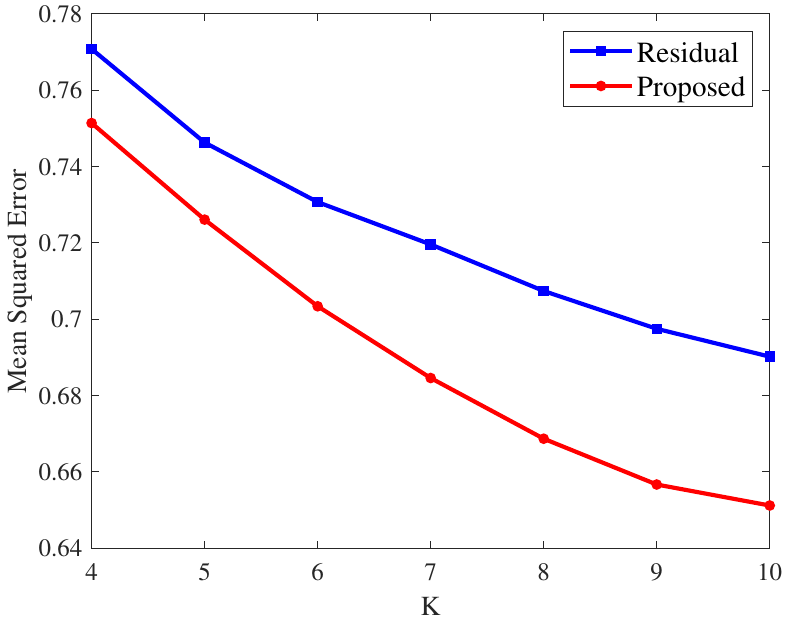}}
\caption{Results of NTK kernel ridge regression of the residual topology and the proposed topology on the Boston house prices dataset. }
\label{NTK_results}
\end{figure}

We use the Boston house prices dataset \cite{harrison1978hedonic} as a testbed that has 13 attributes including the average number of rooms, pupil-teacher ratio, and so on. The task is to predict the house price based on the attributes of a house. The dataset is randomly split into a training set ($90\%$) and a test set ($10\%$).
The mean squared error between predictions and ground truth is computed as the evaluation metric. We vary $K$ from 4 to 10 to make a thorough comparison. The code is written online in Google Colab based on Python neural tangent package (https://github.com/google/neural-tangents). For all $K$, the inference time is no more than 10 seconds. Figure \ref{NTK_results} highlights the consistent improvement of the proposed topology over the residual one. In addition, while the mean squared errors of both models keep going down as $K$ increases, the downward momentum of the proposed topology is stronger. 

\subsection{Generalizability}

Here, we validate the generalizability of the proposed topology with concatenations to see if it can truly deliver competitive results as promised. Suppose that $y_l$ is the output of the $l^{th}$ module in the network of $L$ modules, we characterize the workflow of the proposed topology in the following way:
\begin{equation}
\begin{aligned}
    & y_{l+1} =H_l(y_l),  \\
    & y_{L}  = H_{L-1}(y_0 \oplus y_1 \oplus y_2 \oplus ... y_{L-1}),
\end{aligned}
\end{equation}
where $\oplus$ is a concatenation operator. 
The operator module $H(\cdot)$ can perform multiple operations including batch normalization \cite{ioffe2015batch}, convolution, dropout \cite{srivastava2014dropout}, and so on. While our theoretical analysis revolves around the multiplication operation, it can also scale to the convolution operation because a convolution between two vectors can be re-formulated as matrix multiplication.

The network of the proposed topology is implemented as a drop-in replacement for the DenseNet, which means that the only difference between our network and DenseNet is the shortcut topology. Our model comprises multiple blocks, and each block employs the proposed topology. Like DenseNet, the important hyperparameters for our model are the feature growth rate $k$ and the number of layers in each block. The number of features of a layer in a block is referred to as the growth rate, which regulates the capacity of information passed to the final. We compare the proposed topology with other advanced deep learning benchmark models on the CIFAR-100, Tiny ImageNet, and ImageNet datasets. 

\textbf{CIFAR-100}: We follow the initialization strategy in DenseNet. The DenseNet utilizes stage training: Across the stages, the number of filters is doubled, and the size of feature maps is reduced at the scale of 2. The proposed network includes four blocks. All model configurations for the proposed model follow the protocol in \cite{huang2017densely}. The total number of epochs is 250. The initial learning rate is 0.1 and divided by 10 in every quarter of the total epoch number. We use SGD for training with a weight decay of 0.0001 and a momentum of 0.9. We run each of the proposed models five times and compute the corresponding mean and variance of errors. In Table \ref{densent_compare}, we summarize the experimental results on the CIFAR-100. The network of the proposed topology achieves a slightly higher error rate of $23.52\%$ with much fewer parameters. The proposed model works better at larger growth rates, which is quite different from the DenseNet. Because of the memory constraint, a larger growth rate is prohibitive for the DenseNet. Overall, the proposed topology achieves competitive results over CIFAR-100. 

\begin{table}[htb]
 \centering
\caption{Comparisons of top-1 errors (\%) on CIFAR-100 among the proposed model and other models.  }
 \begin{tabular}{ |p{3.6cm}|p{1.2cm}|c|  }
 \hline
 Network &  Params  & Error($\%$) \\

   \hline
   NIN + Dropout \cite{lin2013network}       & -  & 35.68\\
   \hline
   FractalNet with Dropout \cite{larsson2016fractalnet}  & 38.6M & 35.34 \\
   \hline 
   ResNet (Stocatic Depth) \cite{huang2016deep}   & 1.7M & 37.80 \\
   \hline
   DIANet \cite{huang2020dianet}  &- & 23.02 \\
   \hline
   SpinalNet \cite{kabir2020spinalnet} &- & 35.01 \\
   \hline
   LP-BNN \cite{franchi2020encoding} & - & 23.02 \\
   \hline
   DenseNet (k=12, depth=40)  & 1.0M & 27.55 \\

  DenseNet (k=12, depth=100)  & 7.0M & 23.79 \\

   DenseNet (k=24, depth=100)  & 27.2M & 23.42 \\
  \hline
   Proposed (k=12, depth=40)  & 0.4M & $29.63 \pm 0.017$\\

   Proposed (k=24, depth=40)  &1.3M & $26.21 \pm 0.025$ \\

   Proposed (k=40, depth=40)  &3.6M  & \textbf{23.52 $\pm$ 0.037} \\

%   Proposed (k=64, depth=40)  &8.3M  & 23.30\\

%   \textbf{Proposed (k=64, depth=70)}  &21.2M  &  \textbf{22.76}\\
   \hline 
 \end{tabular}
 \vspace{1ex}
 
 {The errors of compared models are reported by the official implementation. \par}

 \label{densent_compare}
\end{table}

\begin{table}[htb]
 \centering

\caption{Comparisons of top-1 errors (\%) among various advancing models on Tiny ImageNet.}
 \begin{tabular}{ |p{3.8cm}|p{0.6cm}|p{0.8cm}|p{1.7cm}|  }
  \hline
  Network  & l.r. &  Params  & Error($\%$) \\

  \hline
  MobileNetV2 (2018) \cite{sandler2018mobilenetv2}   & 0.1    & 3.5M  & 43.76\\
  \hline
  EfficientNet-B0 (2019) \cite{tan2019efficientnet}  & 0.1    & 5.3M  & 42.91\\
  \hline
  OctResNet50 (2019) \cite{chen2019drop}  & 0.1   & 25.5M  & 47.45\\
   \hline 
  Lambda Network (2020) \cite{bello2020lambdanetworks}  &0.1    & 15.0M  & 58.71\\
   \hline    
  SE-Net (2018) \cite{hu2018squeeze}  &0.05      & 28.1M  & 53.98\\
   \hline   
  Scale-Net (2019) \cite{li2019data}  &0.01      & 31.4M  & 48.59\\
   \hline   
  Ghost-Net (2020) \cite{han2020ghostnet}  & 0.1    & 5.2M & 44.01\\
   \hline   
  RandomWire-WS (2019) \cite{xie2019exploring}  & 0.01   & 31.6M   & 42.11\\
   \hline  
  Proposed A (k=96, depth=41)  & 0.1     & 5.0M  & 42.82$\pm$ 0.31\\
   \hline     
 \textbf{Proposed B (k=108, depth=41)}  & 0.1 & 10.6M  & \textbf{42.04 $\pm$ 0.24}\\
   \hline     
 \end{tabular}
 \vspace{1ex}
 
  {All models are implemented by us. \par}
  
 \label{tinyImageNet}

\end{table}

\textbf{Tiny ImageNet dataset}: This dataset consists of 200 classes with 500 training, 50 validation, and 50 test images per class. The image size is $64\times 64$, which are downsampled from the full images of the ImageNet dataset. In the experiments, we select the following models for comparison: MobileNet-V2 \cite{sandler2018mobilenetv2}, EfficientNet-B0 \cite{tan2019efficientnet}, OctResNet50 \cite{chen2019drop}, Lambda Network \cite{anonymous2021lambdanetworks}, SE-Net \cite{hu2018squeeze}, Scale-Net \cite{li2019data}, Ghost-Net \cite{han2020ghostnet}, and Randomly Wired Network \cite{xie2019exploring}. All these models are well-known new benchmarks. We set the batch size to $64$. We adopt the standard learning rate decay approach. In every 30 epochs, the learning rate is divided by $10$. The initial learning rate is chosen from $\{0.01, 0.05, 0.1\}$. The momentum is 0.9. All models are trained in two TITAN Xp and one GeForce GTX 1080 GPUs. Among all models, it takes at most 742.43 seconds to finish one epoch. Based on our tuning, the appropriate hyperparameters for competitors are shown in Table \ref{tinyImageNet}. We verify two models (k=96, depth=41, init-nf=32 and k=108, depth=41, init-nf=32), each of which consists of three blocks and "init-nf" means the number of features in the first layer of each block. We run the proposed two models five times and compute the mean and variance of errors. Table \ref{tinyImageNet} shows top-1 validation errors of all models, where both proposed models achieve state-of-the-art performance. Particularly, the proposed model at a high growth rate obtains competitive accuracy over all the other models. One notable thing is that given a target performance, the network of the proposed topology uses three times fewer parameters than the randomly wired network.

\textbf{ImageNet dataset}: The ImageNet dataset \cite{deng2009imagenet}
consists of $1.2$ million images for training and $50,000$ images for validation. No other augmentation techniques are employed in our experiments. We follow the basic data augmentation methods, as used in ShuffleNet,
Randomly Wired Network,
DenseNet,
ECANet, and
SENet.
 For model configurations, we follow those of DenseNet \cite{huang2017densely}. We set the batch size as 156, the initial learning rate as 0.1, the weight decay as 0.0001, and the momentum as 0.9. In validation, we adopt the standard 10-crop validation. To be fair, we compare our model with others in the small size regime ($<10M$ parameters) and regular size regime ($\sim 20M$ parameters), respectively. It takes the smaller model around 75 minutes and the larger model around 90 minutes per epoch. We run the larger model three times and the smaller model five times to compute the average and variance of errors. Due to the computational burden of searches, NAS-based models appear in the small regime. Tables \ref{ImageNetResults_small} and \ref{ImageNetResults_regular} highlight the state-of-the-art results achieved by the proposed model. Regarding the small size regime, despite a moderately higher model complexity, the proposed model achieves a performance superior or similar to those of other advanced models. Very favorably, our model is designed based on theoretical analyses. Compared to NAS, our model is free of computationally expensive searches. While for the regular model regime, our model is comparable to other advanced models.

\begin{table}[htb]
 \centering

\caption{The top-1 error (\%) comparisons in small model regime on ImageNet validation set.}
 \begin{tabular}{ |p{3.5cm}|p{1cm}|p{1.5cm}|  }
 \hline
 Network  & params & Error($\%$) \\
  \hline
   MobileNetV2 (2018) \cite{sandler2018mobilenetv2}   & 6.9M & 25.3 \\
   \hline
   ShuffleNet (2018) \cite{zhang2018shufflenet}  & 5.4M & 26.3 \\
   \hline
  NASNet-B (2018) \cite{zoph2018learning}  & 5.3M & 27.2\\
   \hline 
  NASNet-C (2018) \cite{zoph2018learning}  & 4.9M & 27.5\\
   \hline  
  Amoeba-A (2018) \cite{real2019regularized}  & 5.1M & 25.5\\
   \hline     
  Amoeba-B (2018) \cite{real2019regularized}  & 5.3M & 26.0\\
   \hline  
  PNAS (2018) \cite{liu2018progressive}  & 5.1M & 25.8 \\
   \hline  
  DARTS (2019) \cite{liu2018darts}  & 4.9M & 26.9 \\
   \hline 
  FBNet-A (2019) \cite{wu2019fbnet}   & 4.3M & 27.0 \\
  \hline
  RandWire-WS (2019) \cite{xie2019exploring}  & 5.6M & 25.3 $\pm$ 0.25 \\
  \hline
  RegNetX-600MF (2020) \cite{radosavovic2020designing}  & 6.2M & 25.9 $\pm$ 0.03\\
  \hline
  DeiT-Ti (2020) \cite{touvron2021training}  & 5.0M & 25.4  \\
  \hline
  \textbf{Proposed (k=96, depth=45)}  &  9.4M & \textbf{25.2 $\pm$ 0.07}\\
  \hline
   \end{tabular}
   
 \vspace{1ex}
 {The errors of compared models are reported by the official implementation. \par}
 
 \label{ImageNetResults_small}
 \vspace{-2mm}
\end{table}

\begin{table}[htb]
 \centering

\caption{The top-1 error (\%) comparisons in regular model regime on ImageNet validation set.}
 \begin{tabular}{ |p{3.5cm}|p{1cm}|p{1.5cm}|  }
 \hline
 Network  & params & Error($\%$) \\
  \hline
  SENet (2018) \cite{hu2018squeeze} & 26.8M & 23.3 \\
  \hline
  ACNet (2019) \cite{wang2019adaptively} &19.8M & 23.8 \\
  \hline
  DenseNAS-R2 (2020) \cite{fang2020densely}  & 19.5M & 24.2\\
  \hline
  ECA-Net (2020) \cite{wang2020eca} & 24.4M & \textbf{22.5} \\
  \hline 
%   \textbf{Proposed (k=160, depth=41)}   & 22.1M & \textbf{23.5}\\
%   \hline
  \textbf{Proposed (k=180, depth=41)}   & 27.9M & \textbf{22.9 $\pm$ 0.06}\\
  \hline
  \end{tabular}
   
 \vspace{1ex}
 {The errors of compared models are reported by the official implementation. \par}
 
 \label{ImageNetResults_regular}

\end{table}

\subsection{Interpretability}

\begin{figure}[tbh]
\center{\includegraphics[width=\linewidth] {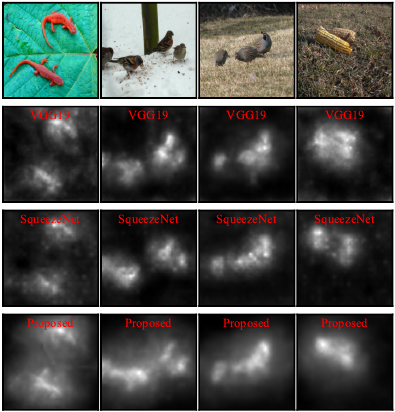}}
\caption{Saliency maps of different models by the FullGrad method. Visually, regarding four images, saliency maps of the proposed model are sharper and their brightest points more conform to the objects.}
\label{saliency}

\end{figure} 

\begin{figure}[tbh]
\center{\includegraphics[width=\linewidth] {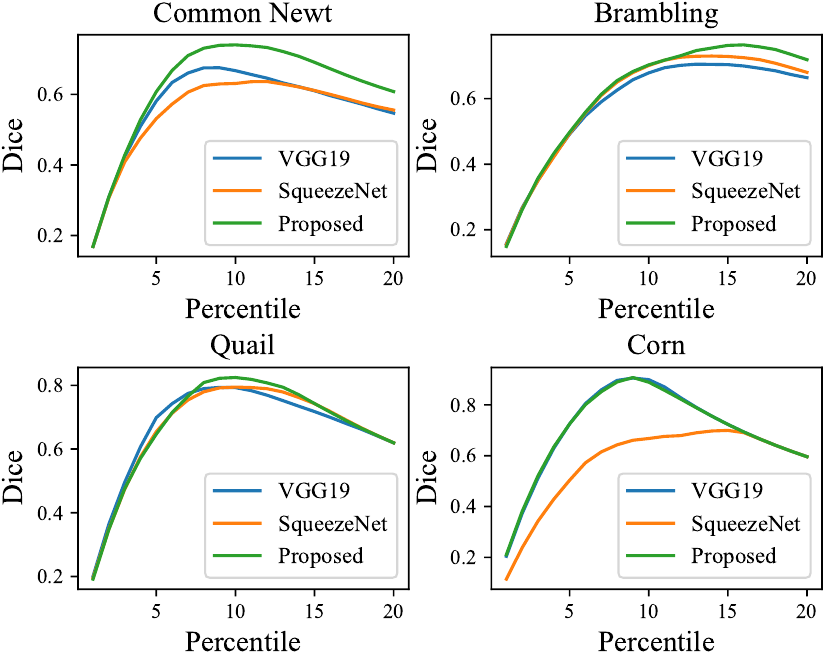}}
\caption{Dice scores between the segmentation of an object and a saliency map as a function of the percentile. The segmentation of a saliency map is obtained by setting the $q$-percentile brightest pixels as one and the rest as zero. }
\label{dicescore}
\vspace{-2mm}
\end{figure} 

Interpretability is a fundamental problem for the development of deep learning \cite{fan2020interpretability, hou2020learning}. Here, we also show the superior interpretability of the proposed model in terms of the saliency map. 

\textbf{Saliency map:} Currently, saliency methods deriving a saliency map by identifying relevance between features and the prediction of a model are the mainstream interpretability methods \cite{arrieta2020explainable}. A myriad of saliency methods are based on gradients \cite{fan2020interpretability}, the idea of which is that the strength of gradients can mirror the extent of how a feature can affect model output. As we know, shortcuts can facilitate training by alleviating gradient explosion and vanishing issues. The mechanism is that shortcuts provide additional paths for straightforward gradient propagation, improving the quality of saliency maps. In the proposed topology, shortcuts directly connect the final layer with all the prior layers, thereby conveying gradients among them. Meanwhile, the proposed topology can generalize, ignoring pixels from the input not located in the object that influences the image label, improving saliency maps as well. Integrating these two aspects, the saliency map of the proposed topology should be more accurate and sharper relative to the network without shortcuts.

We use the FullGrad method \cite{srinivas2019full} to derive saliency maps because it can satisfy two characteristics (dependence and completeness) that the community has deemed important, while other classic methods such as SmoothGrad \cite{smilkov2017smoothgrad}, IntegratedGrad \cite{sundararajan2017axiomatic}, and so on cannot. Dependence describes that a feature is important if it can substantially affect the model output, while completeness is that the individual saliency scores must add up to the model output, which ensures that the total relevance corresponds to the extent of what is detected by a model. We compare our model with classic deep learning models: VGG19 \cite{simonyan2014very} and SqueezeNet \cite{iandola2016squeezenet}. Both models have no shortcuts. We have obtained three our models from three runs in the ImageNet experiments in the regular model regime. VGG19 and SqueezeNet are straightforwardly obtained from the PyTorch library. 

\begin{table}[htb]
 \centering

\caption{The mean Dice scores of different models for 30 images between saliency maps and segmentation. The segmentation is obtained by removing the background from an image.}
 \begin{tabular}{ |c|c|  }
 \hline
   Network & Dice Score \\
  \hline
  VGG19  & $0.609$ \\
  \hline 
  SqueezeNet  & $0.586$ \\
  \hline 
%     ResNet50  & $0.629\pm 0.144$ \\
%   \hline 
%     DenseNet  & $0.623\pm 0.149$ \\
%   \hline 
    \textbf{Proposed($1^{st}$ run)}  & \textbf{0.630} \\
  \hline 
    Proposed($2^{nd}$ run)  & 0.629 \\
  \hline   
    Proposed($3^{rd}$ run)  & 0.627 \\
  \hline 
      Proposed(mean$\pm$std)  & 0.6287$\pm$0.0015 \\
  \hline 
  \end{tabular}
 \label{dice_score}
\end{table}

Saliency maps for four randomly selected ImageNet images from different models are shown in Figure \ref{saliency}. Visually, for all images, the saliency maps of the proposed model are sharper, and the brightest points more tightly conform to the objects, compared to VGG19 and SqueezeNet. In addition, we also quantify the quality of saliency maps. First, we threshold the saliency map by setting the $q$-percentile brightest pixels as one and the rest as zero to get a segmentation map. Then, we use the Dice score ($\frac{2|X \cap Y|}{X\cup Y}$) \cite{dice1945measures} between the segmentation of an object and a saliency map to measure their similarity. This metric by and large can reflect the sharpness and accuracy of a saliency map. The higher the score is, the better interpretability a model has. The obtained segmentation and saliency maps are put in Part C of supplementary materials for conciseness. Figure \ref{dicescore} shows the Dice scores for four objects concerning different percentiles and models. The percentile range is from top-$1\%$ to top-$20\%$ with a step of $1\%$. We find that the proposed model achieves the highest Dice scores over common-newt, brambling, and quail images. For the corn image, the proposed model is comparable to VGG19 but much better than SqueezeNet.

Furthermore, we make a dataset comprising 30 images and their segmentation maps, by randomly selecting images from the ImageNet validation set and manually removing their background. For each pair, we record the maximum Dice score associated with a certain percentile. The mean Dice scores for 30 images are shown in Table \ref{dice_score}. The detailed Dice scores for each image are summarized in Table I of Part D in supplementary materials. There are two highlights from Table \ref{dice_score}. First, the Dice scores of our models surpass those of competitors with a considerable margin, which implies that the saliency maps generated by our model are of higher quality than those of competitors. The second highlight is that our model results are pretty consistent with one another, where the variance among models is only 0.0015. To highlight the improvements made by the proposed model, we conduct the paired t-test between the proposed model and the competitor. The null hypothesis is that the pairwise difference in Dice scores between two models has a mean equal to zero. The test decisions for all pairs are shown in Table II in Appendix C), where all decisions reject the null hypothesis at the default 5\% significance level. This suggests that the improvement by our model is significant.

\section{DISCUSSION}

In \cite{wang2018mixed}, it was demonstrated that the ResNet topology is also intrinsically the densely connected topology. Suppose $X_l=H_l(R_{l-1})$, which is the output of the $l^{th}$ layer, and $R_0=X_0$,
\begin{equation}
    \begin{aligned}
    X_l & = H_l(R_{l-1}) = H_l(H_{l-1}(R_{l-2})+R_{l-2}) \\
        & = H_l(H_{l-1}(R_{l-2})+H_{l-2}(R_{l-3})+R_{l-3}) \\
        & = H_l(\sum_{i=0}^{l-1}H_i(R_{i-1})+R_0) \\
        & = H_l(\sum_{i=0}^{l-1}X_i+X_0) \\
        & = H_l(X_0+X_1+\cdots+X_{l-1}).
    \end{aligned}
\end{equation}
 Therefore, our theoretical results on the densely connected topology can be somehow extended to the ResNet topology.

\begin{figure}[tbh]
\center{\includegraphics[height=2.2in,width=3.2in,scale=0.2] {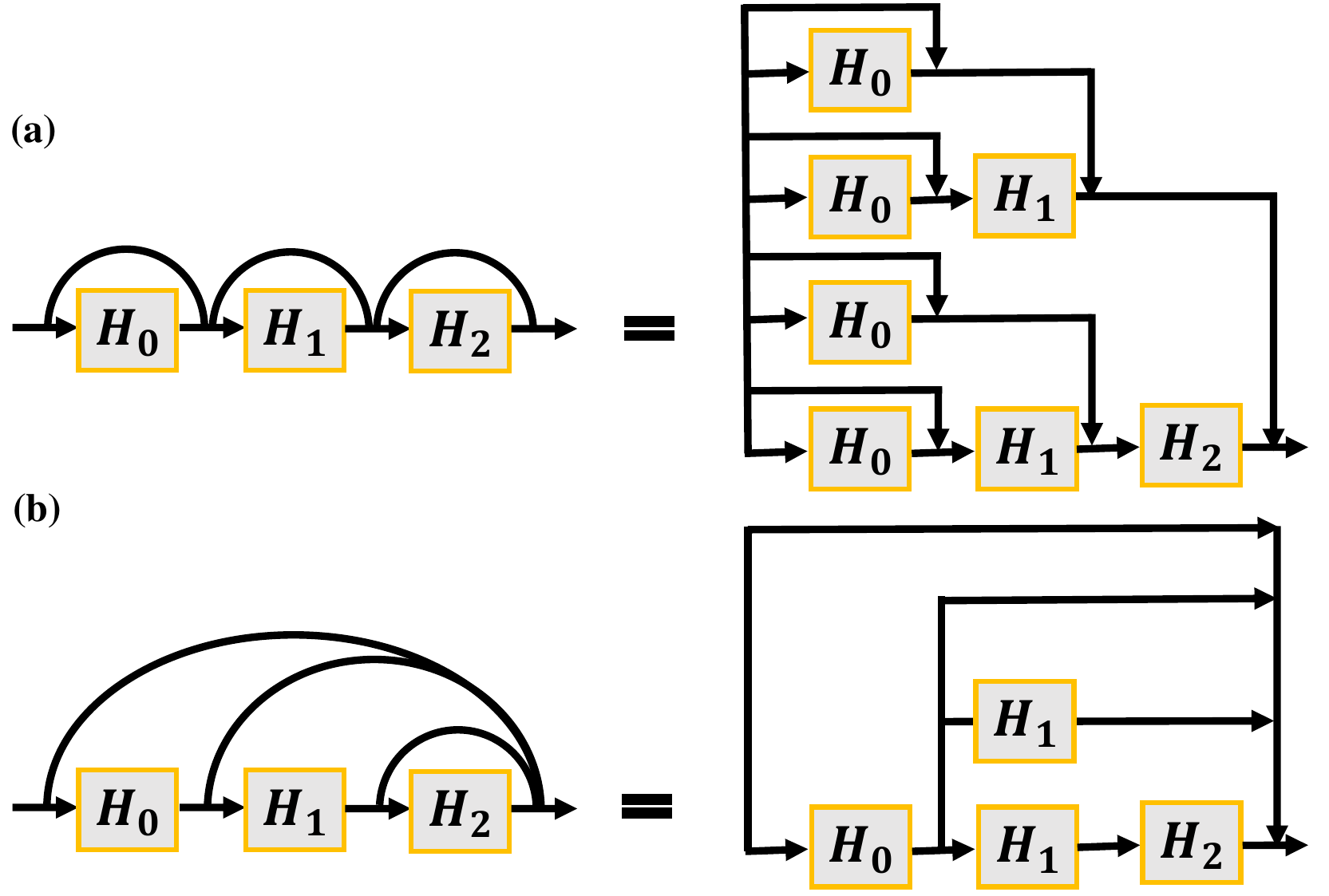}}
\caption{(a) ResNet and its unraveled view; (b) the proposed topology and its unraveled view. The operation at the joints of black lines is summation. }
\label{unraveled_view}
\vspace{-2mm}
\end{figure}

In \cite{veit2016residual}, ResNet is interpreted as an ensemble of many paths of different lengths, and an ablation study shows that deleting a single layer does not affect the performance significantly. In light of ensemble behavior, as shown in Figure \ref{unraveled_view}, given the depth $L$, there are $2^L$ implicit paths connecting the input and output in ResNet, while for the proposed network, the number of implicit paths is $L+1$. Furthermore, in ResNet, every layer has an equal chance of being passed or not passed. However, implicit paths of the proposed topology pass earlier layers more than later layers. For example, in Figure \ref{unraveled_view}(b), only one path connects $H_2$, but three paths connect $H_0$. We conduct the ablation study on the proposed network with k=180 and depth = 41 from Table \ref{ImageNetResults_regular}. We set the outputs of the first and fifth layers of each block as zeros respectively and examine the performance of the network on the test set. Because we have obtained three models from repetitive experiments, the ablation is repeated three times. The results are shown in Table \ref{ensemble}. We can see that undoing the first layer of each block has a significant impact, which causes only 31.84$\%$ accuracy. In contrast, the model with the fifth layer of each block being undone still has the classification accuracy of 61.91$\%$.

\begin{table}[htb]
 \centering

\caption{Performance by undoing different layers in the proposed model to manifest the relative importance of each layer}
 \begin{tabular}{ |c|c|c|c|  }
 \hline
   & Original & Undo Layer 1 & Undo Layer 5\\
  \hline
  Accuracy ($\%$)   & $77.1\pm 0.06$ & $31.84\pm1.81$ & $61.91\pm5.52$\\
  \hline 
%   \textbf{Proposed (k=160, depth=41)}   & 22.1M & \textbf{23.5}\\
%   \hline
  \end{tabular}
   
 \label{ensemble}
 \vspace{-4mm}
\end{table}

\section{CONCLUSION}
In this study, we have theoretically demonstrated the expressivity and generalizability of skip connections in deep learning, with an emphasis on the proposed topology. Then, we have performed comprehensive prediction and classification experiments to corroborate our theoretical findings that the networks of the proposed topology enjoy good expressivity and generalizability. Furthermore, we have also shown that the proposed model embraces improved interpretability in terms of saliency maps and layer importance. We have shared our code and prepared images in \textcolor{blue}{https://github.com/FengleiFan/SparseShortcutTopology}. Future research directions can be put into exploring the utility of network equivalency in neural architecture search studies.

\bibliographystyle{ieeetr}
\bibliography{sample.bib}

\end{document}